\documentclass{new_tlp}

\usepackage{color}
\usepackage{comment}
\usepackage[hidelinks]{hyperref}
\usepackage{amsmath}
\usepackage{xspace}
\usepackage{enumitem}
\usepackage{rotating}
\usepackage{changepage}

\usepackage{todonotes}
\usepackage{color}

\def \daniela#1{\todo[inline,color=yellow!40]{#1}}

\def\AIA{${\cal AIA}$\xspace}

\def\TI{${\cal TI}$\xspace}
\def\kb{${\cal KB}$\xspace}
\def\var#1{{\hbox{{\bf #1\/}}}}
\def\inst#1{{\hbox{{\tt #1\/}}}}
\def\sort#1{{\hbox{{\it #1\/}}}}
\def\action#1{{\hbox{{\it #1\/}}}}
\def\fluent#1{{\hbox{{\it #1\/}}}}
\def\rRM{$RM$\xspace}

\newtheorem{example}{Example}

\definecolor{codegray}{rgb}{0.5,0.5,0.5}

     {\begin{list}{}%
             {\setlength{\leftmargin}{#1}}%
             \item[]%
     }
     {\end{list}}

  \title[]
        {An ASP Methodology for Understanding Narratives about Stereotypical Activities}

  \author[Inclezan et al.]
         {DANIELA INCLEZAN, QINGLIN ZHANG\\
            Miami University, Oxford OH 45056, USA\\
            \email{inclezd,zhangq7@miamioh.edu}
          \and 
          MARCELLO BALDUCCINI\\
            Saint Joseph's University, Philadelphia PA 19131, USA\\
            \email{marcello.balduccini@gmail.com}
          \and 
          ANKUSH ISRANEY\\
            Drexel University, Philadelphia PA 19104, USA\\
            \email{avi26@drexel.edu}}
         
\jdate{March 2003}
\pubyear{2003}
\pagerange{\pageref{firstpage}--\pageref{lastpage}}
\doi{S1471068401001193}

\begin{document}

\maketitle

\begin{abstract}
We describe an application of Answer Set Programming to the understanding of narratives 
about stereotypical activities, demonstrated via question answering.
Substantial work in this direction was done by Erik Mueller, who modeled stereotypical activities 
as {\em scripts}. His systems were able to understand a good number of narratives,
but could not process texts describing exceptional scenarios.
We propose addressing this problem by using a theory of intentions developed by Blount, Gelfond, and Balduccini.
We present a methodology in which we substitute scripts by {\em activities} (i.e., hierarchical 
plans associated with goals) and 
employ the concept of an {\em intentional agent} 
to reason about both normal and exceptional scenarios.
We exemplify the application of this methodology by answering questions about a number of restaurant stories.
This paper is under consideration for acceptance in TPLP.
\end{abstract}

       
\section{Introduction}
\label{intro}
This paper describes an application of Answer Set Programming to the understanding
of narratives. 
According to Schank and Abelson \citeyear{sa77}, stories frequently narrate episodes related 
to {\em stereotypical activities} --- {\em sequences of actions normally
performed in a certain order by one or more actors, according to cultural conventions.}
One example of a stereotypical activity is dining in a restaurant with table service, 
in which the following actions are expected to occur: 
the customer enters, he is greeted by the waiter who leads him to
a table, the customer sits down, reads the menu, orders some dish,
the waiter brings the dish, the customer eats and then asks for the bill, the waiter places
the bill on the table, the customer pays and then leaves.
A story mentioning a stereotypical activity is not required to state explicitly 
all of the actions that are part of it, as it is assumed that the reader is capable
of filling in the blanks with his own commonsense knowledge about the activity \cite{sa77}. 
Consider, for instance, the following narrative:

\begin{example}[Scenario 1, adapted from \cite{m07}]
\label{ex1}
{\em Nicole went to a vegetarian restaurant. She
ordered lentil soup. The waitress set the soup
in the middle of the table. Nicole enjoyed the
soup. She left the restaurant.}
\end{example}

\noindent
Norms indicate, for instance, that customers are expected to pay for their meal.
Readers are supposed to know such conventions, and so
this information is missing from the text.

Schank and Abelson \citeyear{sa77} introduced the concept of a {\em script} to model stereotypical 
activities: a {\em fixed} sequence of actions that are {\em always} executed in a specific
order. Following these ideas, Mueller conducted
substantial work on texts about stereotypical activities:
news about terrorist incidents \citeyear{m04} and restaurant stories \citeyear{m07}.
In the latter he developed a system that took as an input a restaurant story, 
processed it using information extraction techniques, 
and used a commonsense knowledge base about the restaurant domain 
to demonstrate an understanding of the narrative by answering questions
whose answers were not necessarily stated in the text.
The system performed well 
but the rigidity of scripts
did not allow for the correct processing of scenarios 
describing 
exceptions (e.g., someone else paying for the customer's meal). 
To be able to handle such scenarios, 
all 
possible exceptions of a script would have to be foreseen
and encoded as new scripts by the designer of the knowledge base,
which is an important hurdle.

{\em In this paper, we propose a new representation methodology and reasoning approach,
which makes it possible to answer, in both normal and exception scenarios, 
questions about events that did or did not take place. We overcome limitations in
Mueller's work by abandoning the rigid script-based approach. 
To the best of our knowledge, ours is the first scalable approach 
to the understanding of \textbf{exceptional} scenarios.} 

Instead, we propose to view characters in stories about stereotypical activities 
(e.g., the customer, waiter, and cook in a restaurant scenario), as 
BDI agents that {\em intend} to perform some actions in order to achieve certain goals, 
but may not always need to/ be able to perform these actions as soon as intended.
It is instrumental for our purpose to use a {\em theory of intentions}
developed by Blount {\em et al.} 
\citeyear{bgb15}
that introduces the concept of an {\em activity} ---
a sequence of agent actions and sub-activities 
that are supposed to achieve a goal.
The theory of intentions is written in an action language ($\cal{AL}$)
and is translatable
into Answer Set Prolog (ASP) \cite{gl91}. It can be easily coupled with ASP 
commonsense knowledge bases about actions and their effects, and with reasoning algorithms
encoded in ASP or its extensions, to build executable systems. 
Other implementations of BDI agents exist. Some cannot 
be immediately integrated in executable systems \cite{rg91};
it remains to be seen whether others \cite{bhw07} may be more readily integrated in our methodology.

Blount {\em et al.} also introduced an architecture (\AIA) of an {\em intentional agent},
an agent that obeys his intentions. According to \AIA, at each time step, the agent
observes the world, explains observations incompatible with its expectations
(diagnosis), and determines what action to execute next (planning). 
The reasoning module implementing \AIA  
allows an agent to reason about a wide variety of scenarios, including the serendipitous 
achievement of its goal by exogenous actions or the realization that an active activity
has no chance to achieve its goal anymore (i.e., it is futile),
illustrated by the texts below.

\begin{example}[Serendipity]
\label{ex2}
{\em Nicole went to a vegetarian restaurant. She ordered lentil soup.
When the waitress brought her the soup, she told her that it was on the house.
Nicole enjoyed the soup and then left.} 
(The reader should understand that Nicole did not pay for the soup.)
\end{example}

\vspace*{-0.2cm}
\begin{example}[Detecting Futile Activity]
\label{ex4}
{\em Nicole went to a vegetarian restaurant. She sat down and wanted to order lentil soup,
but it was not on the menu.}
(The reader should deduce that Nicole stopped her 
plan of eating lentil soup 
here.)
\end{example}

\vspace*{-0.2cm}
\begin{example}[Diagnosis]
\label{ex6}
{\em Nicole went to a vegetarian restaurant. She
 ordered lentil soup. The waitress brought her a miso soup instead. }
(The reader is supposed to produce some explanations for what may have gone wrong:
either the waitress or the cook misunderstood the order.)
\end{example}

In contrast with \AIA, which encodes an agent's reasoning process about his own goals,
intentions, and ways to achieve them,
we need to represent the reasoning process 
of a (cautious) reader that learns about the
actions of an intentional agent from a narrative. 
For instance, while an intelligent agent creates or selects its own activity to
achieve a goal, in a narrative context, the reader learns about the activity 
that was selected by the agent from the text.
As a consequence, 
{\em one of our goals in this work is to understand what parts of the \AIA 
architecture can be adapted to our purpose of modeling the reasoning of
a narrative reader and how.}
The main difficulty is that stereotypical activities normally include several agents (e.g., customer, waiter, cook), 
not just one.
We had to extend Blount {\em et al.}'s theory to be able to track the 
intentions of several agents at a time.

Our methodology can be applied to narratives about other 
stereotypical activities. 
This is just a first exploration 
of the subject,\footnote{An extended abstract  
on this work was previously published by Inclezan {\em et al.} \citeyear{izbi17};
in previous work by Zhang and Inclezan \citeyear{zi17}, 
only the customer was modeled as a goal-driven agent.
}
which can be further expanded by addressing, for instance,
script variations.
We propose a prototypical implementation of the end-to-end system in which understanding is tested via question answering.
We convert questions to logic forms to closely match their meanings in English and  
encode corresponding rules in ASP to retrieve answers. 
Mueller's collection of training and test excerpts is proprietary, and creating a benchmark of exceptional scenarios
is a laborious task.
As a result, we evaluate our work on a smaller collection of texts collected from the Internet.

In what follows, the paper provides 
a review of related and background work. 
It then continues with the description of the methodology and 
a preliminary implementation, 
and their exemplification on sample scenarios. 
It ends with conclusions and future work.

\section{Related Work}
\label{sec:rel_work}
{\bf Story Understanding.} 
An extensive review of narrative processing systems can be found in Mueller's paper \citeyear{m07}.
Newer systems exist, for example the logic-based systems discussed by Michael \citeyear{lm13} or 
Diakidoy {\em et al.} \citeyear{dkmm15},
but do not focus specifically on stereotypical activities.
The task we undertake here is a more difficult one 
because stories about stereotypical activities 
tend to omit more information about the events taking place
compared to other texts, as such information is expected to be filled in by the reader.

\medskip
\noindent
{\bf Restaurant Narratives.} 
Erik Mueller's work is based on the hypothesis that readers of a text understand it by constructing a mental model 
of the narrative. 
Mueller's system \citeyear{m07} showed an understanding of restaurant
 narratives by answering questions about time and space aspects that were not necessarily mentioned
 explicitly in the text.
His system relied on two important pieces of background knowledge:
(1) a commonsense knowledge base about actions occurring in a restaurant, 
their effects and preconditions, encoded in Event Calculus \cite{s97}
and (2) a script describing a sequence of actions performed by different characters
in a {\em normal} unfolding of a restaurant episode.
The system processed English text using information extraction techniques 
in order to fill out slot values in a template. In particular, it detected the last action 
from the restaurant script that was mentioned in the text, and 
constructed a logic form in which the 
occurrence of all actions in the script up to that last one mentioned was assumed and 
recorded via facts. Clearly, this approach cannot be applied to exceptional cases.
For instance, the last script action identifiable in the scenario in Example \ref{ex2}
is that Nicole left the restaurant. As a result, the reasoning problem constructed by Mueller's system
for this excerpt would state as a fact that Nicole also executed a preceding action in the script, that of 
paying for her meal, which would be incorrect.
Mueller's system was evaluated on 
text excerpts retrieved from the
web or Project Gutenberg collection. 
Scenarios with exceptional cases were not processed correctly because of a lack of flexibility of scripts.

\medskip
\noindent
{\bf Activity Recognition.} The task we undertake here presents some similarities 
to activity recognition, in that it requires observing agents and their environment
in order to complete the picture about the agents' actions and activities.
However, unlike activity recognition, understanding narratives limited to a 
single stereotypical activity (restaurant dining) 
does not require identifying agents' goals, which are always the same for 
each role in our case (e.g., the customer always wants to become satiated).
Gabaldon \citeyear{Gabaldon09} performed activity recognition using a simpler 
theory of intentions by Baral and Gelfond \citeyear{bg05i} that did not consider 
goal-driven agents.
Nieves {\em et al.} \citeyear{ngl13} proposed an argumentation-based approach for
activity recognition, applied to activities defined as pairs of 
a motive and a set of goal-directed actions;
in contrast, in Blount {\em et al.}'s work, basic actions in an activity may, 
but are not required to, have an associated goal.
A few decades earlier, Ng and Mooney \citeyear{nm92} used abduction to create a plan recognition
system and tested it on a collection of short narratives that included restaurant dining. However,
their system cannot reason about serendipitous achievement of an 
agent's goals by someone else's actions (Example \ref{ex2}), nor answer questions about 
an agent's intentions.

\section{Preliminary: Theory of Intentions}
\label{sec:TI}

Blount {\em et al.} \cite{thesisblount13,bgb15} 
developed a theory about the intentions of a goal-driven agent
by substantially elaborating on previous work by Baral and Gelfond \citeyear{bg05i}. 
In their theory, 
each sequence of actions (i.e., plan) of an agent was associated 
with a goal that it was meant to achieve, and the combination of the two
was called an {\em activity}. Activities could have nested sub-activities, and
were encoded using the predicates:
$activity(\var{m})$ ($\var{m}$ is an activity); 
$goal(\var{m}, \var{g})$ (the goal of activity $\var{m}$ is $\var{g}$); 
$length(\var{m}, \var{n})$ (the length of activity $\var{m}$ is $\var{n}$); and
$comp(\var{m}, \var{k}, \var{x})$ (the $\var{k}^{th}$ component of activity 
$\var{m}$ is $\var{x}$, where $\var{x}$ is either an action or a sub-activity).

The authors introduced the concept of an {\em intentional agent}\footnote{
There are similarities between Blount {\em et al.}'s intentional agents and BDI commitment agents \cite{rg91,wm09}.
To the best of our knowledge, they have not yet been studied precisely. However, a link can be drawn, at the intuitive level, as follows. If we consider the perspective that ASP formalizations are typically focused on the {\em beliefs} of an agent about its environment,
an intentional agent can be viewed as an open-minded commitment agent. 
However, if the ASP formalization reflects accurately the physical reality, then it can be viewed as a single-minded commitment agent.
} --- one that 
has goals that it intends to pursue, ``only attempts
to perform those actions that are intended and does so without delay.''
As normally done in our field, 
the agent is expected to possess knowledge about the changing world
around it. This can be represented as a transition diagram 
in which nodes denote {\em physical} states of the world and
arcs are labeled by {\em physically executable} actions that may take the world from one state 
to the other. States describe the values of relevant properties of the world,
where properties are 
divided into fluents (those that can be changed by actions)  
and statics (those that cannot).
To accommodate intentions and decisions of an intentional agent, 
Blount {\em et al.} expanded the traditional transition diagram with 
{\em mental} fluents and actions.
Three important mental fluents 
in their theory are $status(\var{m}, \var{k})$ ($\var{m}$ is in progress if
$\var{k} \geq 0$; not yet started or stopped if $\var{k}=-1$),
$active\_goal($\var{g}$)$ (``goal $\var{g}$ is active''),
and $next\_action(\var{m}, \var{a})$ (``the next physical action to be executed as part of 
activity $\var{m}$ is $\var{a}$''). 
Axioms describe how the execution of physical actions affects the status of activities and sub-activities,
activates (or inactivates) goals and sub-goals, and determines the selection of the next action to be executed 
(see \cite{thesisblount13} for a complete list of axioms).
Mental actions include $select(\var{g})$ and $abandon(\var{g})$ for goals,
and $start(\var{m})$ and $stop(\var{m})$ for activities.
The new transition diagram is encoded in action language $\cal{AL}$; in what follows, we denote
by \TI the ASP translation of the $\cal{AL}$ encoding.

Additionally, Blount {\em et al.} developed an agent architecture \AIA
for an {\em intentional} agent, implemented in CR-Prolog \cite{bg03a,b07} -- an extension of ASP.
Blount \citeyear{thesisblount13} adapted the agent loop proposed by Balduccini and Gelfond \citeyear{bg08}
and outlined the control loop that 
governs the behavior of an intentional agent, which we reproduce in Figure \ref{fig_aia}.

\begin{figure}[hb!]
\fbox{
  \parbox{0.97\textwidth}{
Observe the world and initialize history with observations;
\begin{enumerate}[leftmargin=*,noitemsep,topsep=0pt]
\item[{\bf 1.}] interpret observations;\label{AIA-step-1}
\item[{\bf 2.}] find an intended action $e$; \label{AIA-step-2}
\item[{\bf 3.}] attempt to perform $e$ and \label{AIA-step-3}

update history with a record of the attempt;
\item[{\bf 4.}] observe the world, \label{AIA-step-4}

update history with observations, and

go to step \ref{AIA-step-1}.
\end{enumerate}
}
}
\caption{${\cal AIA}$ control loop}
\label{fig_aia}
\end{figure}

For each step of the control loop, 
we provide a summary of the original description (see pages 43-44 of \cite{thesisblount13}).
In step {\bf 1}, the agent uses diagnostic reasoning to explain unexpected observations,
which involves determining which exogenous (i.e., non-agent) actions may have occurred without
being observed.
From the point of view of our approach, step {\bf 2} is arguably one of the most critical.
The goal of this step is to allow the agent to find an {\em intended action}. 
The following intended actions are considered:
\begin{itemize}[noitemsep,topsep=0pt]
\item
To continue executing an ongoing activity that is expected to achieve its goal;
\item
To stop an ongoing activity whose goal is no longer active (because it has been either achieved, as in Example \ref{ex2}, or abandoned);
\item
To stop an activity that is no longer expected to achieve its goal (as in Example~\ref{ex4}); or
\item
To start a chosen activity that is expected to achieve its goal. 
\end{itemize}
Under certain conditions, there may
be no way for the agent to achieve its goal, or the agent may simply have no goal. In either case,
the agent's intended action is to wait.
For the case when the agent continues executing an ongoing activity,
the fluent $next\_action(\var{m}, \var{a})$ in the theory of intentions 
becomes relevant as it indicates the action in activity $\var{m}$ that the agent would have to attempt next.
In step {\bf 3}, the agent acts and records its attempt to perform the intended action.
In the final step {\bf 4}, the agent observes the values of fluents, the result of his attempt to act from step {\bf 3},
and possibly the occurrence of some exogenous actions.

\smallskip
Restaurant stories require reasoning simultaneously about the intentions of 
multiple goal-driven agents. To accommodate for this, we added an extra argument 
\var{ag} to associate an agent to each mental fluent and action of \TI
(e.g., $status(\var{m}, \var{k})$ became $status(\var{ag}, \var{m}, \var{k})$).
We also extended \TI by the ASP axioms below, needed to make explicit Blount {\em et al.}'s
assumption that an agent has only one top-level goal at a time.
This restriction is important when modeling an external observer and was not fully captured previously by \TI.
The first two axioms say that an agent cannot select a goal
if it already has an active goal or if it selects another goal at the same time.
The third rule says that the stopping of an activity inactivates the goals of all of its 
sub-activities.

\noindent
$
\begin{array}{lll}
impossible(select(Ag, G), I) & \leftarrow & holds(active\_goal(Ag, G_1), I),\ 
                                            possible\_goal(Ag, G).\\
impossible(select(Ag, G), I) & \leftarrow & occurs(select(Ag, G_1), I),\ 
                                            possible\_goal(Ag, G),\\ 
                                            & & G \neq G_1.
\end{array}
$

\noindent
$        
\begin{array}{lll}                                    
\neg holds(active\_goal(Ag, G_1), I+1) & \leftarrow & goal(M_1, G_1), \\
       & &  holds(descendant(Ag, M_1, M), I),\\ 
     & & occurs(stop(Ag, M), I).    
\end{array}
$

\section{Methodology}
\label{sec:method}

In this section, we outline a methodology of using the theory of intentions and
parts of the \AIA architecture to design a program that can show an understanding of 
stories about stereotypical activities, exemplified 
on restaurant stories. 
We distinguish between the story time line containing strictly the events mentioned in the text,
and the reasoning time line corresponding to the mental model that the reader constructs.
We begin with assumptions and the general methodology, on which we elaborate in the next subsections.

\medskip
\noindent
{\bf Assumptions.} 
We assume that a wide coverage commonsense knowledge base (\kb) written in ASP 
is available to us and that it contains information about a large number of 
actions, their effects and preconditions, including actions in the stereotypical activity.
How to actually build such a knowledge base is a difficult research question, but
it is orthogonal to our goal. 
To see the first necessary steps for building such a knowledge base see Diakidoy {\em et al.} \citeyear{dkmm15}.  
In practice, in order to be able to evaluate our methodology, 
we have built a basic knowledge base with core information 
about restaurants and, whenever a scenario needed new information, we expanded the
knowledge base with new actions and fluents. We operated under the assumption 
that all this information would be in \kb from the beginning.
To simplify this first attempt to use a theory of intentions to reason about
stereotypical activities, we assumed that there is only one customer that wants to dine, 
only one waiter, one cook, 
and one ordered dish. 

\medskip
\noindent
{\bf Methodology.} 
According to our methodology, for each input text $t$ and 
set of questions $Q$,
we construct a logic program $\Pi(t, Q)$ (simply $\Pi(t)$ if $Q$ is empty).
Its answer sets represent models of the narrative and answers to questions in $Q$.
This logic program has two parts, one that is pre-defined, and another that depends on the input.

The {\bf pre-defined part} consists of the following items:

\noindent
\fbox{
  \parbox{0.97\textwidth}{
\begin{enumerate}[leftmargin=*,noitemsep,topsep=0pt]
\item The \kb knowledge base, with a core describing sorts, fluents, actions,
and some pre-defined objects 
 relevant to the stereotypical activity of focus;
\item The ASP theory of intentions, \TI;  
\item A module encoding stereotypical activities
as \TI activities for each character;
and
\item A reasoning module, encoding (i) a mapping of time points on the story time line
into points on the reasoning time line; (ii) reasoning components adapted from the \AIA 
architecture to reflect a reader's reasoning process and
expected to allow reasoning about serendipitous achievement of goals,
decisions to stop futile activities, and diagnosis; and 
(iii) a question answering component.
\end{enumerate}
  }
}

\smallskip
The {\bf input-dependent part} (i.e., the logic form obtained by translating the English 
text $t$ and questions in $Q$ into ASP facts) consists of the following: 

\noindent
\fbox{
  \parbox{0.97\textwidth}{
\begin{enumerate}[leftmargin=*,noitemsep,topsep=0pt]
\item[5.] Facts defining objects mentioned in the text $t$
as instances of relevant sorts in \kb; 
\item[6.] Observations about the values of fluents and the occurrences of actions at different points on the {\em story} time line;
\item[7.] Default information about the values of fluents in the initial situation; and
\item[8.] Facts encoding each question in $Q$.
\end{enumerate}
}
}

\subsection{The Core of the Commonsense Knowledge Base \kb}\label{sec:core-KB}
The core of \kb defines knowledge related to the restaurant environment.
It includes a hierarchy of sorts with main sorts \sort{person}, \sort{thing},
\sort{restaurant}, and \sort{location};
\sort{person} has sub-sorts \sort{customer}, \sort{waiter}, and \sort{cook};
and \sort{thing} has sub-sorts \sort{food}, \sort{menu}, and \sort{bill}.
In this paper, the following pre-defined instances of sorts are used:
\inst{entrance}, \inst{kt} (kitchen), \inst{ct} (counter), \inst{outside}, \inst{t} (table) 
are instances of \sort{location};  
\inst{m} is a \sort{menu}; and \inst{b} is the customer's bill.
The core describes actions and fluents related to the restaurant environment 
that can be seen in Table \ref{table:af},
in which \inst{t} denotes the table and we use \var{c} for a \sort{customer};
\var{w} for a \sort{waiter};
\var{ck} for a \sort{cook};
\var{f} for a \sort{food};
\var{r} for a \sort{restaurant};
\var{t1} and \var{t2} for \sort{thing}s;
\var{l}, \var{l1} and \var{l2} for \sort{location}s;
\var{p}, \var{p1}, and \var{p2} for \sort{person}s. 
We denote the agent performing each action \var{a} 
by using the static 
$actor(\var{a}, \var{p})$. Each action has a unique actor, except 
$\action{lead\_to}(\var{w}, \var{c}, \inst{t})$
in which both \var{w} and \var{c} (waiter and customer) are 
considered actors.
All fluents are \emph{inertial} (i.e., they normally maintain their previous values unless changed
by an action), except the five on the last column that are \emph{defined-positive} fluents,
i.e., their positive value is completely defined in terms of other fluents; 
otherwise their default value is false.

\begin{table}
\caption{Important actions and fluents in the restaurant-related core of \kb}
\label{table:af}
\begin{minipage}{\textwidth}
\begin{tabular}{ll}
\hline\hline
{\rotatebox[origin=c]{90}{{\bf Actions}}}
&
\begin{oldtabular} { l l l l }
\action{go}(\var{c}, \var{r}) &  \action{sit}(\var{c}) &
  \action{request}(\var{p1}, \var{t}, \var{p2})\ \ & \action{stand\_up}(\var{c}) \\
\action{greet}(\var{w}, \var{c}) &  \action{pick\_up}(\var{p}, \var{t}, \var{l})&   
  \action{prepare}(\var{ck}, \var{f}) & \action{leave}(\var{c})\\
\action{move}(\var{p}, \var{l1}, \var{l2})\ \ & \action{put\_down}(\var{p}, \var{t}, \var{l})\ \ & 
  \action{eat}(\var{c}, \var{f}) & \action{make\_unavailable}(\var{f}, \var{r})\\
\action{lead\_to}(\var{w}, \var{c}, \inst{t})   & \action{order}(\var{c}, \var{f}, \var{w}) &  
  \action{pay}(\var{c}) &  \action{interference}
\end{oldtabular} 
\\
\hline
{\rotatebox[origin=c]{90}{{\bf Fluents}}}
&
\begin{oldtabular}{ l l l l }
\fluent{hungry}(\var{c})                                                        & \fluent{standing\_by}(\var{p}, \var{l})\ \  &
  \fluent{available}(\var{f}, \var{r})          &  \fluent{order\_transmitted}(\var{c})\\
\fluent{open}(\var{r})                                                  & \fluent{sitting}(\var{c})& 
  \fluent{food\_prepared}(\var{ck}, \var{f})\ \ & \fluent{done\_with\_payment}(\var{c})\\
\fluent{at\_loc}(\var{t}, \var{l})\ \  &  \fluent{holding}(\var{p}, \var{t})&
  \fluent{served}(\var{c})                      & \fluent{satiated\_and\_out}(\var{c})\\
\fluent{in}(\var{c}, \var{r})                  &        \fluent{menu\_read}(\var{c})& 
  \fluent{bill\_generated}(\var{c})             & \fluent{served\_and\_billed}(\var{c})         \\
\fluent{welcomed}(\var{c}) \ \         & \fluent{informed}(\var{p1}, \var{t}, \var{p2})\ \  &
  \fluent{paid}(\inst{b})                       & \fluent{done\_with\_request}(\var{ck}, \var{w})
  
\end{oldtabular}\\
\hline\hline
\end{tabular}
\end{minipage}
\end{table}

Axioms about the direct, indirect effects and preconditions of actions 
are encoded in ASP using standard methods \cite{gk14}. 
We show here the encoding of a direct effect of 
action \action{eat(\var{c}, \var{f})} 
(the food is no longer on the table)
and a condition that renders its execution impossible
(a customer cannot eat unless the food is on the table):

\smallskip
$
\begin{array}{lll}
\neg holds(at\_l(F,\inst{t}),I+1) & \leftarrow & occurs(eat(C,F),I).
\end{array}
$

$
\begin{array} {lll}                       
impossible(eat(C, F), I) & \leftarrow & \neg holds(at\_l(F, \inst{t}), I), customer(C).
\end{array}
$

\noindent
Defined fluents like \fluent{satiated\_and\_out} 
are defined by rules like:

$
\begin{array}{lll}
holds(satiated\_and\_out(C), I) & \leftarrow & holds(satiated(C), I), \ 
holds(at\_l(C, outside), I).
\end{array}
$

The knowledge base \kb contains an exogenous action
(i.e., a non-agent action) called \action{interference} (see Table \ref{table:af}). 
The simultaneous occurrence of this action with  
\action{order}(\var{c}, \var{f}, \var{w}) 
or
\action{request}(\var{p1}, \var{t}, \var{p2})
causes miscommunication, meaning that the food order or general request is not transmitted 
correctly.
We encode this as a non-deterministic direct effect 
in the rules:

\smallskip\smallskip
$
\begin{array}{lll}
holds(informed(W,F,C),I+1) & \leftarrow & occurs(order(C,F,W),I),\\
& &  {\mbox not\ }\neg holds(informed(W,F,C),I).
\end{array}
$

$
\begin{array}{l}
1 \{holds(informed(W,F_1,C), I+1) \ : \ other\_food(F_1, F)\} 1 \ \leftarrow \\  
\ \ \ \ \ \ \ \ \ \ \ \ \ \ \ \ \ \ \ \ \ \ \ \ \ \ \ \ \ \ \ \ \ 
occurs(order(C,F,W),I), \ occurs(interference, I).
\end{array}
$

$
\begin{array}{lll}
other\_food(F_1, F) & \leftarrow & food(F),\ food(F_1),\ F \neq F_1.
\end{array}
$

$
\begin{array}{lll}
\neg holds(informed(W,F_1,C),I+1) & \leftarrow & holds(informed(W,F_2,C),I),\ F_1 \neq F_2.
\end{array}
$

\noindent
The first rule indicates that {\em normally} the waiter 
understands the customer's order correctly.
An exception is when an \action{interference} occurs at the same time as
a customer's action of ordering some food $F$, which causes the waiter 
to understand that the customer is asking for a different food than $F$
(second axiom above).
The fourth rule says that \fluent{informed} is a functional fluent.
The \kb contains similar rules for the action representing all other communication acts, \action{request}(\var{p1}, \var{t}, \var{p2}). 

\subsection{Encoding Stereotypical Activities}
\label{act_seq}

Stories about stereotypical activities include multiple characters
with their own goals and actions. We modeled these as \TI activities
stored in the \kb, as shown in Table~\ref{table:act}
(recall that \inst{t}, \inst{m}, \inst{b}, \inst{kt}, and \inst{ct} stand for the table, menu, 
bill, kitchen, and counter respectively).

For the {\bf customer}, we defined 
an activity $c\_act(C, R, W, F)$ 
that should be read as ``customer $C$
goes to restaurant $R$ where he communicates to waiter $W$ an order for food $F$.''
We found that modeling the customer's activity as a nested one with sub-activities 
allowed reasoning about a larger number of exceptional scenarios
compared to its formalization as a flat activity.
We introduced sub-activities $c\_subact\_1(C, F, W)$  --
``$C$ consults the menu and communicates an order for food $F$ to $W$,''
and $c\_subact\_2(C, W)$ -- 
``$C$ asks $W$ for the bill and pays for it.''
We encoded a {\bf waiter}'s activities via objects of the form $w\_act(W, C, F_1, F_2)$ --  
``waiter $W$ {\em understood} that customer $C$ ordered food $F_1$ and 
served food $F_2$ to him.'' 
To allow reasoning about different types of exceptions including miscommunication, 
$F_1$ may be different from $F_2$ and/or the food actually ordered by $C$.
We named the {\bf cook}'s activities $ck\_act(Ck, F, W)$ -- 
``cook $Ck$ prepares food $F$ for waiter $W$.''
Food $F$ may not be the one ordered by the customer nor the one requested
by $W$. We encoded activities in Table~\ref{table:act} via ASP rules like:

\noindent
$
\begin{array}{l}
\ \ activity(c\_act(C, R, W, F)) \leftarrow customer(C), restaurant(R), waiter(W), food(F).
\end{array}
$

\noindent
$
\begin{array}{lll}
\ \ comp(c\_act(C, R, W, F),\ 1,\ go(C, R)) & \leftarrow & activity(c\_act(C, R, W, F)).
\end{array}
$

\vspace{-0.2cm}
\noindent
$\ \ \ \ \dots$
\vspace{-0.1cm}

\noindent
$
\begin{array}{lll}
\ \ comp(c\_act(C, R, W, F),\ 9,\ leave(C)) & \leftarrow & activity(c\_act(C, R, W, F)).
\end{array}
$

\noindent
$
\begin{array}{lll}
\ \ length(c\_act(C, R, W, F),\ 9) & \leftarrow & activity(c\_act(C, R, W, F)).
\end{array}
$

\noindent
$
\begin{array}{l}
\ \ goal(c\_act(C, R, W, F),\ satiated\_and\_out(C)) \leftarrow activity(c\_act(C, R, W, F)).
\end{array}
$

\noindent
and specified the actor of each activity using the predicate 
$actor(\var{m}, \var{p})$.

\noindent
\begin{table}
\caption{Activities (and sub-activities) for the main roles in a restaurant story}
\label{table:act}
\begin{minipage}{\textwidth}
\begin{tabular}{rl}
\hline\hline
{\rotatebox[origin=c]{90}{{\bf Customer}}}
&
\begin{oldtabular}{l}
\begin{oldtabular}{ l l }
{\bf Name:} & \boldmath$c\_act(C, R, W, F)$ \\
{\bf Plan:} & $[\ go(C, R),\ lead\_to(W, C, \inst{t}),\ sit(C),\ 
          c\_subact\_1(C, F, W),\ eat(C, F),\ \ $\\ 
   & $\ \ c\_subact\_2(C, W), \ stand\_up(C),\ move(C, \inst{t}, \inst{entrance}),\ leave(C)\ ]$\\
{\bf Goal:}     & $satiated\_and\_out(C)$   
\end{oldtabular}
\\
\begin{oldtabular}{ l l}
\ \ \ \ &
\begin{oldtabular}{l l r}
\hline
Name:      & $c\_subact\_1(C, F, W)$ & (sub-activity)\\
Plan:      & $[\ pick\_up(C, \inst{m}, \inst{t}),\ put\_down(C, \inst{m}, \inst{t}),\ 
                 order(C, F, W)\ ]$ &\\
Goal:      & $order\_transmitted(C)$ &
\\ 
\hline
Name:  & $c\_subact\_2(C, W)$  & (sub-activity)\\
Plan:  & $[\ request(C, \inst{b}, W),\ pay(C, \inst{b})\ ]$ &\\
Goal:  & $done\_with\_payment(C)$ &
\end{oldtabular}
\end{oldtabular}
\end{oldtabular}
\\ 
\hline
{\rotatebox[origin=c]{90}{{\bf Waiter}}}
&
\begin{oldtabular}{ l l }
{\bf Name:} & \boldmath{$w\_act(W, C, F_1, F_2)$} \\
{\bf Plan:} & $[\ greet(W, C), \ lead\_to(W, C, \inst{t}),\ 
        move(W, \inst{t}, \inst{kt}),\ 
        request(W, F_1, \inst{ck}),$\\
  & $\ \ pick\_up(W, F_2, \inst{kt}),\ move(W, \inst{kt}, \inst{t}),\ 
        put\_down(W, F_2, \inst{t}),\ move(W, \inst{t}, \inst{ct}),$\\
  & $\ \ pick\_up(W, \inst{b}, \inst{ct}),\ 
        move(W, \inst{ct}, \inst{t}),\ put\_down(W, \inst{b}, \inst{t})\ ]$\\
{\bf Goal:} & $served\_and\_billed(C)$
\end{oldtabular}
\\
\hline
{\rotatebox[origin=c]{90}{{\bf Cook}}}
&
\begin{oldtabular}{ l l }
{\bf Name:} & \boldmath$ck\_act(Ck, F, W)$ \\
{\bf Plan:} & $[\ prepare(Ck, F, W) \ ]$\\
{\bf Goal:} & $done\_with\_request(Ck, W)$
\end{oldtabular}
\\
\hline\hline
\end{tabular}
\vspace{-1\baselineskip}
\end{minipage}
\end{table} 

\normalsize
\noindent
{\bf Default Information about Restaurant Scenarios.}
In relation to characters' activities,
we also define some default facts that complement the information that can be extracted from
a narrative. These state that, in any restaurant scenario, the customer initially 
(i.e., at time step 0 on the reasoning time line) selects 
the goal of being $satiated\_and\_out$. 
Similarly, the waiter selects its goal $served\_and\_billed(C)$
as soon as customer $C$ arrives, and the cook selects its goal
$done\_with\_request(Ck, W)$ immediately after receiving a food request from waiter $W$.
This is encoded as follows:

$
\begin{array}{lll}
occurs(select(C, satiated\_and\_out(C)), 0) & \leftarrow & customer(C).
\end{array}
$

$
\begin{array}{l}
occurs(select(W, served\_and\_billed(C)), I+1) \leftarrow waiter(W),\\ 
\ \ \ \ \ \ holds(arrived(C, R), I+1), 
                \ \neg holds(arrived(C, R), I).
\end{array}
$

$
\begin{array}{l}
occurs(select(Ck, done\_with\_request(Ck, W)), I+1) \leftarrow cook(Ck),\\
\ \ \ \ \ \ holds(informed(Ck, F, W), I+1),
              \ \neg holds(informed(Ck, F, W), I).
\end{array}
$

\noindent
Once a goal is selected, the program assumes that each character 
starts one of their candidate activities that can achieve
this goal. The assumption is formalized by:

$
\begin{array}{l}        
1\{occurs(start(P, M), I+1) : actor(M, P), goal(M, G)\}1 \leftarrow 
occurs(select(P, G), I).\\  
\end{array}
$

\subsection{Reasoning Module}
\label{rm}
We now concentrate on the reasoning module of the \AIA architecture associated
with \TI, and determine what parts of it can be imported/ adapted and what new rules
need to be added in order to capture the reasoning process of a reader of a narrative.
In what follows, we denote the reader's reasoning module by \rRM
and start with a few key points in its \AIA-inspired construction: 

\noindent
$\bullet$ A reader needs to map observations about fluents and actions (i.e., a {\em history})
into
the predicates $holds$ and $occurs$ used for reasoning.
It also needs 
to perform diagnosis (just like an intentional agent would)
when reading pieces of information that are unexpected, 
such as a waiter bringing the wrong dish. 
Thus {\em the temporal projection and diagnostic modules of \AIA are imported into \rRM.}

\noindent
$\bullet$ The reader needs to fill the story time line with new time points 
(and thus construct what we call a {\em reasoning time line}) 
to accommodate mental and physical actions not mentioned in the text.
{\em This is achieved by adding new rules to \rRM}, 
in which we denote story vs. reasoning time steps
by predicates $story\_step$ and $step$, respectively, and introduce predicate
$map(\var{s}, \var{i})$ to say that story step \var{s} is mapped into reasoning step \var{i}:

$
\begin{array}{lll}
1 \{map(S, I) : step(I) \} 1 & \leftarrow & story\_step(S).
\end{array}
$

$
\begin{array}{lll}
\neg map(S, I) & \leftarrow &   map(S_1, I_1),\ S < S_1,\ I \geq I_1,\ story\_step(S),\ step(I).
\end{array}
$

\noindent
Information about story steps that need to be mapped into consecutive reasoning time steps,
extracted from the input in the form of $next\_st(\var{s}, \var{s1})$ facts, is encoded by the rule:

$
\begin{array}{lll}
map(S_1, I+1) & \leftarrow & next\_st(S, S_1),\ map(S, I).
\end{array}
$

\noindent
Observations about the occurrence of actions and values of fluents recorded
from the text using predicates $st\_hpd$ and $st\_obs$ as described in 
Subsection \ref{lf} are translated
into observations on the reasoning time line via the rules:

$
\begin{array}{lll}
hpd(A, V, I) & \leftarrow & st\_hpd(A, V, S),\ map(S, I).\\
obs(F, V, I) & \leftarrow & st\_obs(F, V, S),\ map(S, I).
\end{array}
$

\noindent
Finally, we do not want to create time steps on the reasoning time line
when no action is believed to have occurred (i.e., gaps). We encode this using the rules 

$
\begin{array}{l}
smtg\_occurs(I) \ \leftarrow \ occurs(A, I).\\
\ \leftarrow \ last\_assigned(I),\ step(J),\ J < I,\ \mbox{not }smtg\_occurs(J).
\end{array}
$

\noindent
where $last\_assigned(\var{i})$ is true if \var{i} is the last time step
on the reasoning time line that has a correspondent on the story time line:

$
\begin{array}{lll}
\neg last\_assigned(I) \leftarrow map(S, J),\ step(I),\ step(J),\ I < J.\\
last\_assigned(I) \leftarrow map(S, I),\ \mbox{not } \neg last\_assigned(I).
\end{array}
$

\def\query#1{{\hbox{{\it #1\/}}}}

\noindent
$\bullet$ A reader may be asked questions about the story. We support yes/no, when, who, and where questions related to events. A question is represented by an atom $question(q)$, where $q$ is a term encoding the question, e.g., $occur(A)$ (``did action $A$ occur?''), $when(A)$ (``when did $A$ occur?''). Answers are encoded by atoms $answer(q, a)$, where $a$ is the answer. For example, $answer(occur(pay(nicole,b)),yes)$ states that the answer to question ``Did Nicole pay the bill?'' is yes.
Rules are used for identifying and returning the answers. Due to space limitations, we only briefly illustrate the rules for $question(occur(A))$. 
Suppose we want to reason about the occurrence of a specific event in a story. In that case, a positive answer is returned if the reader has definite reasons for believing that the event happened. This is encoded by the rule:

$
\begin{array}{lll}
answer(occur(A),\ yes) & \leftarrow & question(occur(A)),\ physical{\_} action(A),\ step(I),\\ & & occurs(A, I).
\end{array}
$

\noindent
Answering a definite ``no" is less straightforward, as it requires ensuring that the action did not happen at \emph{any} step: 

$
\begin{array}{l}
maybe(A) \leftarrow physical{\_}action(A),\ step(I),\ \mbox{not } \neg occurs(A,\ I). \\
answer(occur(A),\ no) \leftarrow question(occur(A)), physical{\_}action(A),\ step(I),\\
\hspace*{1.59in}\mbox{not } answer(occur(A),\ yes),\ \mbox{not } maybe(A).
\end{array}
$

\noindent
The first rule states that $A$ \emph{may have occurred} if, for some step $I$, there is lack of evidence that $I$ did not occur. The second rule yields answer ``no'' if there is no evidence that the action definitely occurred and no reason to believe that the action may have occurred.

\smallskip
Blount {\em et al.} considered four categories of histories 
and described the 
encoding of the corresponding agent strategies in \AIA. 
We now analyze each category separately and discuss its suitability for 
\rRM, as well as pertinent changes:

{\bf 1. No goal nor activity to commit to.} {\em In \AIA, the agent waits.}
In stereotypical activities, agents have active pre-defined goals/ activities,
so this is not relevant to \rRM.

\smallskip
{\bf 2. A top-level activity is active but its goal is not. }
{\em In \AIA, the agent stops the activity.}
Serendipitous achievement of a character's goal by someone else's actions
can happen in narratives about stereotypical activities (see Example \ref{ex2}).
Thus \AIA histories of category 2 are relevant and the 
corresponding rules from \AIA are included in \rRM.

\smallskip
{\bf 3. A top-level activity and its goal are active.}
{\em In \AIA, the agent performs the next action,
unless there are no chances for the activity to still achieve its goal
(i.e., the activity is deemed {\em futile}), in which case 
it is stopped.}
\rRM imports the corresponding rules for the agent strategy, 
but changes the definition of a futile activity, since \rRM captures the reasoning
of an external observer, not that of the agent involved in the activity. 
We add default background knowledge about situations that render an activity
futile to the module describing activities and sequences (see Subsection \ref{act_seq}). 
For example, we encode the information that 
an activity of type $c\_act(\var{c}, \var{r}, \var{w}, \var{f})$ is futile if \var{r} 
is observed to be closed when \var{c} wants to enter; 
if \var{f} is observed to be unavailable at \var{r}; etc., via rules like:

$
\begin{array}{lll}
futile(c\_act(C, R, W, F), I) & \leftarrow & obs(open(R), false, I), activity(c\_act(C, R, W, F)).
\end{array}
$

$
\begin{array}{lll}
futile(c\_act(C, R, W, F), I)  & \leftarrow & obs(available(F, R), false, I),\\ & & activity(c\_act(C, R, W, F)).
\end{array}
$

\noindent
More complex rules can also be added, such as that this activity is futile if
no table is available and the customer is impatient.

\smallskip
{\bf 4. A goal is active but there is no active activity to achieve it.}
{\em The agent needs to find a plan (i.e., start a new activity).}
Our cautious reader is not expected 
to guess or assume the new plan that a character 
computes or selects, but rather determine that re-planning is required.
We introduce a new mental action \action{replan}(\var{g}) where \var{g} is a goal,
and add the following rule to \rRM:

$
\begin{array}{lll}
occurs(replan(Ag, G),I) & \leftarrow & categ\_4\_hist(G,I), \mbox{not }impossible(replan(Ag, G), I),\\
     & & \mbox{not } futile(Ag, G, I).
\end{array}
$

\smallskip
The complete set of \rRM rules for reasoning about activities and 
categories of histories is shown in
Figure \ref{fig:res}. 

\begin{figure}[htb!]
\figrule
\begin{center}
\begin{oldtabular}{ll}
  \begin{oldtabular}{r}
  \tiny{ 1} \\ 
  \tiny{ 2} \\ 
  \tiny{ 3} \\
  \tiny{ 4} \\
  \tiny{ 5} \\
  \\
  \tiny{ 6} \\
  \\
  \tiny{ 7} \\
  \tiny{ 8} \\
  \tiny{ 9} \\
  \tiny{10} \\
  \\
  \tiny{11} \\
  \tiny{12} \\
  \tiny{13}\\
  \tiny{14}\\
  \\
  \tiny{15}\\
  \tiny{16}\\
  \tiny{17}\\
  \\
  \tiny{18}\\
  \tiny{19}
  \end{oldtabular}
  &
  \begin{oldtabular}{l}
  $
  \begin{array}{lll}
  categ\_2\_hist(Ag, M, I) & \leftarrow & \neg holds(minor(Ag, M), I),\ holds(active(Ag, M), I),\\   
  & & goal(M, G),\ \neg holds(active\_goal(Ag, G), I).\\
  categ\_3\_hist(Ag, M, I) & \leftarrow & \neg holds(minor(Ag, M), I),\ holds(in\_progress(Ag, M), I).\\
  categ\_4\_hist(Ag, G, I) & \leftarrow & \neg holds(minor(Ag, G), I),\ holds(active\_goal(Ag, G),I),\\
  & & \neg holds(in\_progress(Ag, G),I).
  \end{array}
  $\\   
        \\

  $
  \begin{array}{lll}
  occurs(stop(Ag, M),I) & \leftarrow & categ\_2\_hist(Ag, M,I).
  \end{array}
  $\\
        \\
        
  $
  \begin{array}{lll}
  occurs(AA,I) & \leftarrow & categ\_3\_hist(Ag, M, I),\ \mbox{not } futile(Ag, M, I),\\
  & & \neg holds(minor(Ag, M), I),\     holds(in\_progress(Ag, M), I),\\ 
  & & holds(next\_action(Ag, M, AA), I),\ \mbox{not } impossible(AA, I).
  \end{array}$\\        

  $
  \begin{array}{lll}
  occurs(stop(Ag, M), I) & \leftarrow & categ\_3\_hist(Ag, M, I),\ futile(Ag, M, I),\ 
        activity(M).
        \end{array}
        $\\
        \\
        
        $
        \begin{array}{lll}
        mental\_inertial\_fluent(replanned(Ag, G)) & \leftarrow & agent(Ag), possible\_goal(Ag, G).
        \end{array}
        $\\
        
        $
  \begin{array}{lll}
        holds(replanned(Ag, G), I+1) & \leftarrow & occurs(replan(Ag, G), I).
        \end{array}
        $\\
        
        $
        \begin{array}{lll}
        impossible(replan(Ag, G), I)  & \leftarrow & holds(replanned(Ag, G), I).\\
        impossible(replan(Ag, G), I)  & \leftarrow & occurs(start(Ag, M), I),\ possible\_goal(Ag, G).
        \end{array}
        $\\
        \\
        
  $
  \begin{array}{lll}
        occurs(replan(Ag, G), I) & \leftarrow & categ\_4\_hist(Ag, G, I),\ 
                                                \mbox{not }impossible(replan(Ag, G), I),\\
                                 &            & \mbox{not }futile(Ag, G, I).     
        \end{array}
        $\\
        
        $
  \begin{array}{lll}
   occurs(wait(Ag), I) & \leftarrow & categ\_4\_hist(Ag, G, I), futile(Ag, G, I).  
   \end{array}
   $\\
   \\
   
   $
   \begin{array}{lll}
   justified(Ag, A, I) & \leftarrow & holds(next\_action(Ag, M, A), I), physical\_agent\_action(Ag, A).
   \end{array}
   $\\
   
   $
   \begin{array}{ll}
   \leftarrow & occurs(A, I), \mbox{not }justified(Ag, A, I), physical\_agent\_action(Ag, A).
   \end{array}
   $
   \end{oldtabular}
\end{oldtabular} 
\end{center}
\figrule
\caption{\rRM rules for reasoning about activities and histories.}\label{fig:res}
\end{figure}

This concludes the description of the main parts of \rRM and that of
the pre-defined parts in the logic program constructed according to our methodology.

\subsection{Logic Form}
\label{lf}
Now, we describe the input-dependent part of the logic program that is constructed for
every narrative.
The text is translated into a logic form that contains two parts:
(a) definitions of instances of relevant sorts, e.g. \sort{customer}, 
\sort{waiter}, \sort{food}, \sort{restaurant}; and
(b) observations about the values of fluents and occurrence of actions 
in relation to the {\em story} time line. While this may be accomplished in various ways, currently we adopt an approach that comes from a combination of techniques introduced by Lierler {\em et al.}~\citeyear{LierlerIG17} and Balduccini {\em et al.} \citeyear{bb06}. In the former, the text is translated into a Discourse Representation Structure (DRS) \cite{kampreyle93} by integrating the outputs of the \textsc{lth} semantic role labeler and of \textsc{coreNLP}. While the DRS is a logical representation, it is still largely focused on linguistic elements rather than on concepts relevant to an action theory. 
Balduccini {\em et al.} \citeyear{bb06} bridged this gap by means of special ASP rules that carry out a final mapping step into an Object Semantic Representation (OSR) featuring the elements listed earlier in this paragraph.  

As an example, let us illustrate how the sentence ``Nicole went to a  vegetarian restaurant'' is translated to our target logic form. Following Lierler {\em et al.}'s approach \citeyear{LierlerIG17}, \textsc{lth} is used to identify entities and verbs and to label entities by the roles they play in the verbs as identified by the \textsc{PropBank}\footnote{The \textsc{PropBank} project \cite{propbank} 
provides a corpus of text annotated with information about basic semantic propositions.} 
frame schemas. This results in 
$$\emph{[V (go.01) went] [A1 Nicole] [A4 ``to a vegetarian restaurant'']}$$
where \emph{go.01} denotes sense $1$ of verb \emph{go} from the Ontonotes Sense Groupings, \emph{A1} is the verb's \emph{entity in motion}, and \emph{A4} is its \emph{end point}. Next, \textsc{coreNLP} carries out mention detection and coreference resolution, grouping the phrases from the output of \textsc{lth} that denote the same object. Then, a postprocessing step occurs, which assigns unique labels to the  the entities of the DRS, yielding the first two rows of the DRS from Figure \ref{fig:DRS}. This step also leverages the \textsc{PropBank} frame schemas to generate the description of the events of the DRS found in the third row of Figure \ref{fig:DRS}. Finally, the postprocessing step assigns time steps to the verbs according to their syntactic ordering in the passage, resulting in the final row of the Figure.
\begin{figure}[tbp]
\fbox{
\begin{tabular}{c}
$r1$,$r2$, $e1$\\
\hline
$entity(r1)$, $entity(r2)$, $property(r1,nicole)$, $property(r2,``vegetarian\ restaurant'')$\\
$event(e1)$, $eventType(e1,go\_01)$, $eventArgs(e1,a1,r1)$, $eventArgs(e1,a4,r2)$\\
\end{tabular}
}
\caption{DRS for the sample sentence}\label{fig:DRS}
\end{figure}
\emph{OSR rules} from \cite{bb06} are then used to generate the final representation of the story. While at this stage the OSR rules are created manually, it is conceivable that they can be automatically generated from the \textsc{PropBank} frame schemas. A sample rule that maps verb \emph{go.01} to the occurrence of an action is:
\[
\begin{array}{l}
st\_hpd(go(Actor,Dest),true,S) \leftarrow\\
\hspace*{1in}event(Ev), eventType(Ev,go\_01),\\
\hspace*{1in}eventArgs(Ev,a1,EActor), property(EActor,Actor), \\
\hspace*{1in}eventArgs(Ev,a4,EDest), property(EDest,Dest).
\end{array}
\]
The ORS rules can also introduce new constants (e.g., \inst{cook1}) when the name of one of the entities is not given in the text.
Note that, in order to distinguish between the story time line and the reasoning time line,
we substitute the predicates $obs(\var{f}, \var{v}, \var{i})$ and 
$hpd(\var{a}, \var{v}, \var{i})$ normally used in the description 
of histories \cite{bgb15} by 
$st\_obs(\var{f}, \var{v}, \var{s})$ 
(fluent \var{f} from the \kb has value \var{v} at time step \var{s}
in the story time line, where \var{v} may be \inst{true} or \inst{false}) and 
$st\_hpd(\var{a}, \var{v}, \var{s})$ 
(action \var{a} from the \kb was observed to have occurred if \var{v} is \inst{true}, or 
not if \var{v} is \inst{false}, at
time step \var{s} in the story time line).
In addition to the observations obtained directly from the text (mentioned there explicitly),
the logic form also contains {\em default}, commonsensical observations such as
the fact that the restaurant is assumed to be open, 
the customer is hungry, the waiter is at the entrance, and so on. 

\begin{example}[Logic Form for a Text]
\label{lf_ex}
The text in Example \ref{ex1} is thus translated into a logic form
that includes the following facts in addition to the default observations:

\smallskip
\small
\begin{tabular}{l l}
$
\begin{array} {l}
customer(\inst{nicole}).\\
restaurant(\inst{veg\_r}).\\
food(\inst{lentil\_soup}).\\
waitress(\inst{waitress}).\\
cook(\inst{cook1}).
\end{array}
$
&
$
\begin{array}{l}
st\_hpd(go(\inst{nicole}, \inst{veg\_r}), \inst{true}, 0).\\
st\_hpd(order(\inst{nicole}, \inst{lentil\_soup}, \inst{waitress}), \inst{true}, 1).\\
st\_hpd(put\_down(\inst{waitress}, \inst{lentil\_soup}, \inst{t}), \inst{true}, 2).\\
st\_hpd(eat(\inst{nicole}, \inst{lentil\_soup}), \inst{true}, 3).\\
st\_hpd(leave(\inst{nicole}), \inst{true}, 4).
\end{array}
$
\end{tabular}
\normalsize
\end{example}

Questions are translated, in a similar way, to a logic form consisting of atoms of the form $question(q)$, discussed earlier. For instance, questions \ref{q1} and \ref{q2} below are linked to actions $leave(\inst{nicole})$ and 
$pay(\inst{nicole},\inst{b})$ from the \kb, resp., yielding the logic forms:
\vspace{-0.15cm}
\begin{align}
   \text{{\em Did Nicole leave the restaurant?} \ \ \ \ Logic form: $question(occur(leave(\inst{nicole})))$}\tag{q1}\label{q1}\\
   \text{{\em Did Nicole pay for the soup?} \ \ \ Logic form: $question(occur(pay(\inst{nicole},\inst{b})))$}\tag{q2}\label{q2}
\end{align} 

\section{Methodology Application}
\label{eval}
We applied our methodology to a collection of stories 
describing normal scenarios and different types of exceptional scenarios.
In this section, we exemplify our outcomes on a few illustrative stories.
We test understanding via questions about the occurrence of an event,
which may not be explicitly mentioned in the text.
Such questions are problematic for systems based on statistical methods.
Given that we use a logic approach, based on ASP and its extensions,
the correctness of the answers provided by our methodology can be proven
for certain classes of text-question pairs via methods similar to those employed by Todorova and Gelfond \citeyear{tg12}.
In the future, we plan to perform a more formal evaluation on a larger corpus,
possibly based on recent story benchmarks 
(e.g., \cite{rbr13,mvyka16}).
Mueller's corpora are proprietary and thus not available.

\subsection{Normal Scenario in Example \ref{ex1}} 
The answer set of the program $\Pi(\ref{ex1})$ obtained according to our methodology
contains the $occurs(\var{a}, \var{i})$ atoms shown in \ref{app2},
where \var{a} is an action and \var{i} is a time point on the reasoning time line.
We use this case as a baseline when explaining the output of exceptional scenarios.
As expected, the customer's actions are interleaved with those of the
waitress and cook. As well, an agent's mental actions 
(i.e., selecting a goal and starting/ stopping an activity) 
take one time step when no other action of the same agent occur. 

\subsection{Serendipitous Achievement of Goal in Example \ref{ex2}}
The logic form for this scenario is identical to the one for Example \ref{ex1} 
shown in Example~\ref{lf_ex}, except that the three observations about actions
taking place at story time points 2--4 are replaced by

\small
$
\begin{array}{l}
st\_hpd(pay(\inst{owner},\inst{b}), \inst{true}, 2).\ \ 
st\_hpd(put\_down(\inst{waitress}, \inst{lentil\_soup}, \inst{t}), \inst{true}, 3).\\
st\_hpd(eat(\inst{nicole},\inst{lentil\_soup}), \inst{true}, 4).\ \ 
st\_hpd(leave(\inst{nicole}), \inst{true}, 5).
\end{array}
$
\normalsize

\noindent
where \inst{owner} is a new instance of sort \sort{person}.
Program $\Pi(\ref{ex2})$ has several answer sets, each one mapping the occurrence of 
action $pay(\inst{owner},\inst{b})$ into a different reasoning time step from 12 to 18.
Each of these answer sets also contains similar $occurs$ atoms to $\Pi(\ref{ex1})$ 
up to time step 20 when Nicole eats the soup and the waitress stops her activity (given that
the bill is already paid). 
From then on, it contains the following $occurs$ predicates:

\smallskip
\small

$
\begin{array}{l}
occurs(stop(\inst{waitress},w\_act(\inst{waitress},\inst{nicole},\inst{lentil\_soup},\inst{lentil\_soup})),20)
\end{array}
$

$
\begin{array}{lll}
occurs(eat(\inst{nicole},\inst{lentil\_soup}),20) & \ \ & occurs(stand\_up(\inst{nicole}),23)\\
occurs(start(\inst{nicole},c\_subact\_2(\inst{nicole})),21) & & occurs(move(\inst{nicole},\inst{t},\inst{entrance}),24)\\
occurs(stop(\inst{nicole},c\_subact\_2(\inst{nicole})),22) & & occurs(leave(\inst{nicole}),25)
\end{array}
$

$
\begin{array}{l}
occurs(stop(\inst{nicole},c\_act(\inst{nicole},\inst{veg\_r},\inst{lentil\_soup})),26)
\end{array}
$
\normalsize
\smallskip

\noindent
Thus, our system understands that Nicole has stopped $c\_subact\_2$ 
immediately after starting it because she realized that its goal is already fulfilled.
Note that approaches that do not view characters as goal-driven agents
(including \cite{Gabaldon09,nm92,m07}) face substantial difficulties or simply cannot handle 
serendipitous scenarios.
All answer sets of program $\Pi(\ref{ex2}) \cup \{\ref{q1}, \ref{q2}\}$ contain the additional 
atoms 

\small
$
\begin{array}{l}
answer(occur(leave(nicole)),yes)\ \ \ \ \ \ 
answer(occur(pay(nicole,b)),no)
\end{array}
$
\normalsize

\noindent
specifying that the answer to question
\ref{q1} is {\em Yes}, while the answer to \ref{q2} is {\em No}.

\subsection{Futile Activity in Example \ref{ex4}}
The logic form for Example \ref{ex4} contains the following observations: 

\smallskip
\small
$
\begin{array}{l}
st\_hpd(go(\inst{nicole}, \inst{veg\_r}), \inst{true}, 0).\ \ 
st\_hpd(sit(\inst{nicole}), \inst{true}, 1).\\ 
st\_hpd(pick\_up(\inst{nicole}, \inst{m}, \inst{t}), true, 2).\ \ 
st\_obs(available(\inst{lentil\_soup}, \inst{veg\_r}), false, 3).
\end{array}
$
\normalsize

\smallskip
\noindent
and a fact $next\_st(2,3)$ stating that story time steps 2 and 3 should be translated
into consecutive reasoning time steps (see Section \ref{rm}).

Program $\Pi(\ref{ex4})$ has multiple answer sets varying in the explanation for $st\_obs$:
exogenous action $make\_unavailable(\inst{lentil\_soup}, \inst{veg\_r})$ occurs at a different
reasoning time points in each of them. 
Otherwise, these answer sets contain the same $occurs$ predicates
as $\Pi(\ref{ex1})$ until step 9; then Nicole stops her futile activity
and replans for her still active goal:

\smallskip
\small
$
\begin{array}{l}
occurs(stop(\inst{nicole}, c\_act(\inst{nicole},\inst{veg\_r},\inst{waitress}, \inst{lentil\_soup})),10)\\
occurs(replan(\inst{nicole},satiated\_and\_out(\inst{nicole})),11)
\end{array}
$
\normalsize

\smallskip
\noindent
Thus the reader is cautious and does not make any assumptions about
Nicole leaving the restaurant. 
As expected, it does not state that Nicole ate lentil soup either, which would be impossible,
nor that she paid for anything.


\subsection{Diagnosis in Example \ref{ex6}}
The logic form for Example \ref{ex6} contains a new instance of sort \sort{food}, \inst{miso\_soup}, and 
the observations:

\smallskip
\small
$
\begin{array}{l}
st\_hpd(go(\inst{nicole}, \inst{veg\_r}), \inst{true}, 0).\\
st\_hpd(order(\inst{nicole}, \inst{lentil\_soup}, \inst{waitress}), \inst{true}, 1).\\ 
st\_hpd(put\_down(\inst{waitress}, \inst{miso\_soup}, \inst{t}), \inst{true}, 2).
\end{array}
$
\normalsize

\noindent
The program $\Pi(\ref{ex6})$ has two answer sets, containing explanations on what went
wrong. 

\smallskip\noindent
{\em Answer Set 1.} The first answer set 
indicates that the waitress started a different activity at time step 4
than the one in the normal scenario in Example \ref{ex1}:

\smallskip
$
\begin{array}{l}
occurs(start(\inst{waitress},w\_act(\inst{waitress},\inst{nicole},\inst{miso\_soup},\inst{miso\_soup})),4)
\end{array}
$

\smallskip\noindent
This activity can be read as ``the waitress {\em understood} that Nicole ordered a miso soup and
served her a miso soup.'' Recall that a waiter's possible activities are of the form
$w\_act(W, C, F_1, F_2)$, where $F_1$ may be a different food than the one actually 
ordered by the customer $C$, which allows reasoning about miscommunication between customer 
and waiter as in this answer set. Also, $F_2$ may be a different food than $F_1$, meaning
that the food served may not be the one that the waiter recorded 
as being ordered by the customer, which we will see in the second answer set.
Our program considers all possible activities that can satisfy an agent's goal and decides 
that the activity that can explain the later enfolding of the scenario must be the activity that the 
agent is actually executing.

This first answer set contains the same $occurs$ predicates as $\Pi(\ref{ex1})$
up to time step 10, and then:

\smallskip
\small
$
\begin{array}{l}
occurs(interference,11)\\ 
occurs(order(\inst{nicole},\inst{lentil\_soup},\inst{waitress}),11) \\
occurs(move(\inst{waitress},\inst{t},\inst{kt}),12)\\ 
occurs(stop(\inst{nicole},c\_subact\_1(\inst{nicole},\inst{lentil\_soup})),12)\\
occurs(request(\inst{waitress},{\bf miso\_soup},\inst{cook1}),13) \\
occurs(select(\inst{cook1},done\_with\_request(\inst{cook1},\inst{waitress})),14)\\
occurs(start(\inst{cook1},ck\_act(\inst{cook1},\inst{miso\_soup},\inst{waitress})),15)\\
occurs(prepare(\inst{cook1},{\bf miso\_soup},\inst{waitress}),16) \\
occurs(stop(\inst{cook1},ck\_act(\inst{cook1},\inst{miso\_soup},\inst{waitress})),17)\\ 
occurs(pick\_up(\inst{waitress},{\bf miso\_soup},\inst{kt}),17) \\
occurs(move(\inst{waitress},\inst{kt},\inst{t}),18) \\
occurs(put\_down(\inst{waitress},{\bf miso\_soup},\inst{t}),19)\\
\end{array}
$
\normalsize

\smallskip
\noindent
Here, the reader's explanation is that there was some interference at time step 11
when Nicole ordered the lentil soup. As a result, the waitress misunderstood the order
to be for miso soup and the cook followed her request. 

\smallskip\noindent
{\em Answer Set 2.}
The second answer set differs from the first in terms of what occurs at time steps 
11-13, in particular

\small
$
\begin{array}{l}
occurs(interference,13)\\
occurs(request(\inst{waitress},\inst{lentil\_soup},\inst{cook1}),13)
\end{array}
$
\normalsize

\noindent
and the activities started by the waiter and cook at time steps 4 and 15:

\small
$
\begin{array}{l}
occurs(start(\inst{waitress},w\_act(\inst{waitress},\inst{nicole},\inst{lentil\_soup},{\bf miso\_soup})),4)\\
occurs(start(\inst{cook1},ck\_act(\inst{cook1}, {\bf miso\_soup},\inst{waitress})),15)
\end{array}
$
\normalsize

\noindent
It corresponds to a second possible explanation in which the waitress understood the
order correctly and the misunderstanding/ interference occurred at step 13 when she
communicated the order to the cook.
(Note that the reader makes no assumptions about what happens next.)

\smallskip\noindent
{\em Question Answering.}
These two answer sets show how reasoning by cases in ASP is useful to
answering questions like: 
{\em Did the waitress ask the cook to prepare a lentil soup?}
whose answer would be {\em No} in the first case (answer set 1) 
and {\em Yes} in the second one (answer set 2).
In the future, we envision answering questions like
{\em Why did Nicole receive a wrong order?}
by producing the answers
{\em A1: Because the waitress misunderstood the order.}/ 
{\em A2: Because the cook misunderstood the order.}

\subsection{Discussion}
While we exemplified and tested our methodology on restaurant scenarios, 
our approach is equally applicable to other stereotypical activities,
if we maintain the assumption that the knowledge base contains the 
relevant commonsense information. 
The main task when addressing a new stereotypical activity is defining
the different \TI activities, including goals, of each involved agent. 
Part of this process can be automated by starting from a 
rigid and centralized script learned in an unsupervised manner 
(e.g., \cite{rkp10})
and then assigning its actions to the activities of different agents, 
those performing them.
Determining (sub-)goals and splitting activities into sub-activities 
is a more challenging problem, which deserves substantial attention.

\section{Conclusions and Future Work}
In this work, we proposed a new methodology for automating the 
understanding of narratives about stereotypical activities. 
As a first main contribution, we overcame limitations of Mueller's work \citeyear{m07} 
related to exceptional scenarios. 
To achieve this, we had to abandon Mueller's rigid script-based approach and use instead 
a substantial portion of a state-of-the-art intelligent agent architecture, 
augmented with support for intentions.
Our second contribution is extending the architecture of an intentional agent, 
\AIA \cite{bgb15}, to model a third-person observer.
We exemplified our methodology on several types of scenarios.
In the future, we intend to create an extensive story corpus, which will be a laborious task,
to further test our methodology.

\bibliographystyle{acmtrans}
\bibliography{restaurant_stories}

\begin{thebibliography}{}

\bibitem[\protect\citeauthoryear{Balduccini}{Balduccini}{2007}]{b07}
{\sc Balduccini, M.} 2007.
\newblock {CR-MODELS}: An inference engine for {CR}-{P}rolog.
\newblock In {\em Proceedings of {LPNMR} 2007}, {C.~Baral}, {G.~Brewka}, {and}
  {J.~S. Schlipf}, Eds. LNCS, vol. 4483. Springer, 18--30.

\bibitem[\protect\citeauthoryear{Balduccini, Baral, and Lierler}{Balduccini
  et~al\mbox{.}}{2007}]{bb06}
{\sc Balduccini, M.}, {\sc Baral, C.}, {\sc and} {\sc Lierler, Y.} 2007.
\newblock {\em {H}andbook of {K}nowledge {R}epresentation}.
\newblock Foundations of Artificial Intelligence. Elsevier, Chapter 20.
  {K}nowledge {R}epresentation and {Q}uestion {A}nswering.

\bibitem[\protect\citeauthoryear{Balduccini and Gelfond}{Balduccini and
  Gelfond}{2003}]{bg03a}
{\sc Balduccini, M.} {\sc and} {\sc Gelfond, M.} 2003.
\newblock {L}ogic {P}rograms with {C}onsistency-{R}estoring {R}ules.
\newblock In {\em Proceedings of Commonsense-03}. {AAAI} {P}ress, 9--18.

\bibitem[\protect\citeauthoryear{Balduccini and Gelfond}{Balduccini and
  Gelfond}{2008}]{bg08}
{\sc Balduccini, M.} {\sc and} {\sc Gelfond, M.} 2008.
\newblock The {AAA} architecture: An overview.
\newblock In {\em Architectures for Intelligent Theory-Based Agents, Papers
  from the 2008 {AAAI} Spring Symposium, 2008}. {AAAI} {P}ress, 1--6.

\bibitem[\protect\citeauthoryear{Baral and Gelfond}{Baral and
  Gelfond}{2005}]{bg05i}
{\sc Baral, C.} {\sc and} {\sc Gelfond, M.} 2005.
\newblock Reasoning about {I}ntended {A}ctions.
\newblock In {\em Proceedings of AAAI-05}. AAAI Press, 689--694.

\bibitem[\protect\citeauthoryear{Blount}{Blount}{2013}]{thesisblount13}
{\sc Blount, J.} 2013.
\newblock An {A}rchitecture for {I}ntentional {A}gents.
\newblock Ph.D. thesis, Texas Tech University, Lubbock, TX, USA.

\bibitem[\protect\citeauthoryear{Blount, Gelfond, and Balduccini}{Blount
  et~al\mbox{.}}{2015}]{bgb15}
{\sc Blount, J.}, {\sc Gelfond, M.}, {\sc and} {\sc Balduccini, M.} 2015.
\newblock A theory of intentions for intelligent agents.
\newblock In {\em Proceedings of {LPNMR} 2015}, {F.~Calimeri}, {G.~Ianni},
  {and} {M.~Truszczynski}, Eds. LNCS, vol. 9345. Springer, 134--142.

\bibitem[\protect\citeauthoryear{Bordini, H{\"{u}}bner, and Wooldridge}{Bordini
  et~al\mbox{.}}{2007}]{bhw07}
{\sc Bordini, R.~H.}, {\sc H{\"{u}}bner, J.~F.}, {\sc and} {\sc Wooldridge, M.}
  2007.
\newblock {\em Programming Multi-Agent Systems in {A}gent{S}peak Using
  {J}ason}.
\newblock John Wiley \& Sons, Ltd.

\bibitem[\protect\citeauthoryear{Diakidoy, Kakas, Michael, and Miller}{Diakidoy
  et~al\mbox{.}}{2015}]{dkmm15}
{\sc Diakidoy, I.-A.}, {\sc Kakas, A.}, {\sc Michael, L.}, {\sc and} {\sc
  Miller, R.} 2015.
\newblock Star: A system of argumentation for story comprehension and beyond.
\newblock 12th International Symposium on Logical Formalizations of Commonsense
  Reasoning (Commonsense'15). 64--70.

\bibitem[\protect\citeauthoryear{Gabaldon}{Gabaldon}{2009}]{Gabaldon09}
{\sc Gabaldon, A.} 2009.
\newblock Activity recognition with intended actions.
\newblock In {\em Proceedings of {IJCAI} 2009}, {C.~Boutilier}, Ed. 1696--1701.

\bibitem[\protect\citeauthoryear{Gelfond and Kahl}{Gelfond and
  Kahl}{2014}]{gk14}
{\sc Gelfond, M.} {\sc and} {\sc Kahl, Y.} 2014.
\newblock {\em Knowledge Representation, Reasoning, and the Design of
  Intelligent Agents}.
\newblock Cambridge University Press.

\bibitem[\protect\citeauthoryear{Gelfond and Lifschitz}{Gelfond and
  Lifschitz}{1991}]{gl91}
{\sc Gelfond, M.} {\sc and} {\sc Lifschitz, V.} 1991.
\newblock {C}lassical {N}egation in {L}ogic {P}rograms and {D}isjunctive
  {D}atabases.
\newblock {\em New Generation Computing\/}~{\em 9,\/}~3/4, 365--386.

\bibitem[\protect\citeauthoryear{Inclezan, Zhang, Balduccini, and
  Israney}{Inclezan et~al\mbox{.}}{2017}]{izbi17}
{\sc Inclezan, D.}, {\sc Zhang, Q.}, {\sc Balduccini, M.}, {\sc and} {\sc
  Israney, A.} 2017.
\newblock Understanding restaurant stories using an {ASP} theory of intentions
  ({E}xtended abstract).
\newblock In {\em Technical Communications of the 33rd International Conference
  on Logic Programming (ICLP-TC 2017)}. OASIcs.

\bibitem[\protect\citeauthoryear{Kamp and Reyle}{Kamp and
  Reyle}{1993}]{kampreyle93}
{\sc Kamp, H.} {\sc and} {\sc Reyle, U.} 1993.
\newblock {\em From discourse to logic}. Vol. 1,2.
\newblock Kluwer.

\bibitem[\protect\citeauthoryear{Lierler, Inclezan, and Gelfond}{Lierler
  et~al\mbox{.}}{2017}]{LierlerIG17}
{\sc Lierler, Y.}, {\sc Inclezan, D.}, {\sc and} {\sc Gelfond, M.} 2017.
\newblock Action languages and question answering.
\newblock In {\em {IWCS} 2017 - 12th International Conference on Computational
  Semantics - Short papers}.

\bibitem[\protect\citeauthoryear{Michael}{Michael}{2013}]{lm13}
{\sc Michael, L.} 2013.
\newblock Story understanding... calculemus!
\newblock 11th International Symposium on Logical Formalizations of Commonsense
  Reasoning (Commonsense'13). 64--70.

\bibitem[\protect\citeauthoryear{Mostafazadeh, Vanderwende, Yih, Kohli, and
  Allen}{Mostafazadeh et~al\mbox{.}}{2016}]{mvyka16}
{\sc Mostafazadeh, N.}, {\sc Vanderwende, L.}, {\sc Yih, W.-t.}, {\sc Kohli,
  P.}, {\sc and} {\sc Allen, J.} 2016.
\newblock Story cloze evaluator: Vector space representation evaluation by
  predicting what happens next.
\newblock In {\em Proceedings of RepEval'16}. Association for Computational
  Linguistics, 24--29.

\bibitem[\protect\citeauthoryear{Mueller}{Mueller}{2004}]{m04}
{\sc Mueller, E.~T.} 2004.
\newblock Understanding script-based stories using commonsense reasoning.
\newblock {\em Cognitive Systems Research\/}~{\em 5,\/}~4, 307--340.

\bibitem[\protect\citeauthoryear{Mueller}{Mueller}{2007}]{m07}
{\sc Mueller, E.~T.} 2007.
\newblock Modelling space and time in narratives about restaurants.
\newblock {\em Literary and Linguistic Computing\/}~{\em 22,\/}~1, 67--84.

\bibitem[\protect\citeauthoryear{Ng and Mooney}{Ng and Mooney}{1992}]{nm92}
{\sc Ng, H.~T.} {\sc and} {\sc Mooney, R.~J.} 1992.
\newblock Abductive plan recognition and diagnosis: {A} comprehensive empirical
  evaluation.
\newblock In {\em Proceedings of the 3rd International Conference on Principles
  of Knowledge Representation and Reasoning (KR'92)}. 499--508.

\bibitem[\protect\citeauthoryear{Nieves, Guerrero, and Lindgren}{Nieves
  et~al\mbox{.}}{2013}]{ngl13}
{\sc Nieves, J.~C.}, {\sc Guerrero, E.}, {\sc and} {\sc Lindgren, H.} 2013.
\newblock Reasoning about human activities: an argumentative approach.
\newblock In {\em Twelfth Scandinavian Conference on Artificial Intelligence,
  {SCAI} 2013, Aalborg, Denmark, November 20-22, 2013}. 195--204.

\bibitem[\protect\citeauthoryear{Palmer, Gildea, and Kingsbury}{Palmer
  et~al\mbox{.}}{2005}]{propbank}
{\sc Palmer, M.}, {\sc Gildea, D.}, {\sc and} {\sc Kingsbury, P.} 2005.
\newblock The {P}roposition {B}ank: An annotated corpus of semantic roles.
\newblock {\em Computational Linguistics\/}~{\em 31,\/}~1 (Mar.), 71--106.

\bibitem[\protect\citeauthoryear{Rao and Georgeff}{Rao and
  Georgeff}{1991}]{rg91}
{\sc Rao, A.~S.} {\sc and} {\sc Georgeff, M.~P.} 1991.
\newblock Modeling rational agents within a {BDI}-architecture.
\newblock In {\em Proceedings of the 2nd International Conference on Principles
  of Knowledge Representation and Reasoning (KR'91). Cambridge, MA, USA, April
  22-25, 1991.} 473--484.

\bibitem[\protect\citeauthoryear{Regneri, Koller, and Pinkal}{Regneri
  et~al\mbox{.}}{2010}]{rkp10}
{\sc Regneri, M.}, {\sc Koller, A.}, {\sc and} {\sc Pinkal, M.} 2010.
\newblock Learning script knowledge with web experiments.
\newblock In {\em Proceedings of ACL '10}. 979--988.

\bibitem[\protect\citeauthoryear{Richardson, Burges, and Renshaw}{Richardson
  et~al\mbox{.}}{2013}]{rbr13}
{\sc Richardson, M.}, {\sc Burges, C. J.~C.}, {\sc and} {\sc Renshaw, E.} 2013.
\newblock Mctest: A challenge dataset for the open-domain machine comprehension
  of text.
\newblock In {\em EMNLP}. ACL, 193–--203.

\bibitem[\protect\citeauthoryear{Schank and Abelson}{Schank and
  Abelson}{1977}]{sa77}
{\sc Schank, R.~C.} {\sc and} {\sc Abelson, R.~P.} 1977.
\newblock {\em Scripts, Plans, Goals, and Understanding: An Inquiry into Human
  Knowledge Structures}.
\newblock Lawrence Erlbaum.

\bibitem[\protect\citeauthoryear{Shanahan}{Shanahan}{1997}]{s97}
{\sc Shanahan, M.} 1997.
\newblock {\em Solving the Frame Problem}.
\newblock MIT Press.

\bibitem[\protect\citeauthoryear{Todorova and Gelfond}{Todorova and
  Gelfond}{2012}]{tg12}
{\sc Todorova, Y.} {\sc and} {\sc Gelfond, M.} 2012.
\newblock Toward {Q}uestion {A}nswering in {T}ravel {D}omains.
\newblock In {\em Correct Reasoning}. 311--326.

\bibitem[\protect\citeauthoryear{Wooldridge}{Wooldridge}{2009}]{wm09}
{\sc Wooldridge, M.} 2009.
\newblock {\em An Introduction to MultiAgent Systems\/}, 2nd ed.
\newblock Wiley Publishing.

\bibitem[\protect\citeauthoryear{Zhang and Inclezan}{Zhang and
  Inclezan}{2017}]{zi17}
{\sc Zhang, Q.} {\sc and} {\sc Inclezan, D.} 2017.
\newblock An application of {ASP} theories of intentions to understanding
  restaurant scenarios.
\newblock In {\em Proceedings of PAoASP'17}.

\end{thebibliography}

\newpage
\appendix

\newpage
\section{Partial Output for the Normal Scenario in Example \ref{ex1}}
\label{app2}
$
\begin{array}{l}

map(0, 2)\\
map(1, 11)\\
map(2, 19)\\
map(3, 20)\\
map(4, 31)\\

occurs(select(\inst{nicole}, satiated\_and\_out(\inst{nicole})),0) \\
occurs(start(\inst{nicole}, c\_act(\inst{nicole},\inst{veg\_r},\inst{waitress}, \inst{lentil\_soup})),1) \\
occurs(go(\inst{nicole},\inst{veg\_r}),2) \\
occurs(select(\inst{waitress},served\_and\_billed(\inst{nicole})),3)\\
occurs(start(\inst{waitress},w\_act(\inst{waitress},\inst{nicole},\inst{lentil\_soup},\inst{lentil\_soup})),4)\\
occurs(greet(\inst{waitress},\inst{nicole}),5) \\
occurs(lead\_to(\inst{waitress},\inst{nicole},\inst{t}),6) \\
occurs(sit(\inst{nicole}),7) \\
occurs(start(\inst{nicole}), c\_subact\_1(\inst{nicole},\inst{lentil\_soup}, \inst{waitress})),8) \\
occurs(pick\_up(\inst{nicole},\inst{m},\inst{t}),9) \\
occurs(put\_down(\inst{nicole},\inst{m},\inst{t}),10) \\
occurs(order(\inst{nicole},\inst{lentil\_soup},\inst{waitress}),11) \\
occurs(stop(\inst{nicole},c\_subact\_1(\inst{nicole},\inst{lentil\_soup},\inst{waitress})),12)\\
occurs(move(\inst{waitress},\inst{t},\inst{kt}),12) \\
occurs(request(\inst{waitress},\inst{lentil\_soup},\inst{cook1}),13) \\
occurs(select(\inst{cook1},done\_with\_request(\inst{cook1},\inst{waitress})),14)\\
occurs(start(\inst{cook1},ck\_act(\inst{cook1},\inst{lentil\_soup},\inst{waitress})),15)\\
occurs(prepare(cook1,\inst{lentil\_soup},\inst{waitress}),16) \\
occurs(stop(\inst{cook1},ck\_act(\inst{cook1},\inst{lentil\_soup},\inst{waitress})),17)\\
occurs(pick\_up(\inst{waitress},\inst{lentil\_soup},\inst{kt}),17) \\
occurs(move(\inst{waitress},\inst{kt},\inst{t}),18) \\
occurs(put\_down(\inst{waitress},\inst{lentil\_soup},\inst{t}),19) \\
occurs(eat(\inst{nicole},\inst{lentil\_soup}),20)\\
occurs(start(\inst{nicole},c\_subact\_2(\inst{nicole}, \inst{waitress})),21) \\
occurs(request(\inst{nicole},\inst{b},\inst{waitress}),22) \\
occurs(move(\inst{waitress},\inst{t},\inst{ct}),23) \\
occurs(pick\_up(\inst{waitress},\inst{b},\inst{ct}),24) \\
occurs(move(\inst{waitress},\inst{ct},\inst{t}),25) \\
occurs(put\_down(\inst{waitress},\inst{b},\inst{t}),26) \\
occurs(stop(\inst{waitress},w\_act(\inst{waitress},\inst{nicole},\inst{lentil\_soup},\inst{lentil\_soup})),27)\\
occurs(pay(\inst{nicole},\inst{b}),27) \\
occurs(stop(\inst{nicole}, c\_subact\_2(\inst{nicole}, \inst{waitress})),28) \\
occurs(stand\_up(\inst{nicole}),29) \\
occurs(move(\inst{nicole},\inst{t},\inst{entrance}),30)\\
occurs(leave(\inst{nicole}),31) \\
occurs(stop(\inst{nicole}, c\_act(\inst{nicole},\inst{veg\_r},\inst{waitress}, \inst{lentil\_soup})),32)
\end{array}
$

\end{document}